# Multi-Label Phase Diagram Prediction in Complex Alloys via Physics-Informed Graph Attention Networks

Eunjeong Park[1], Amrita Basak[1,*]

[1] Department of Mechanical Engineering, Pennsylvania State University, University Park, PA, 16802, USA

*correspondence: aub1526@psu.edu

**Abstract**
Accurate phase equilibria are foundational to alloy design because they encode the underlying thermodynamics governing stability, transformations, and processing windows. However, while the CALculation of Phase Diagrams (CALPHAD) provides a rigorous thermodynamic framework, exploring multicomponent composition-temperature space remains computationally expensive and is typically limited to sparse section. To enable rapid phase mapping and alloy screening, we propose a physics-informed graph attention network (GAT) that learns element-aware representations and couples them with thermodynamic constraints for multi-label phase-set prediction in the Ag-Bi-Cu-Sn alloy system. Using about 25,000 equilibrium states generated with pycalphad, each composition-temperature point is represented as a four-node element graph with atomic fractions and elemental descriptors as node features. The model combines graph attention, global pooling, and a multilayer perceptron to predict nine relevant phases. To improve physical consistency, we incorporate thermodynamic constraints, including a Gibbs phase rule-based cap on phase multiplicity, local smoothness away from phase boundaries, and pure-phase feasibility, applied as training penalties or as an inference-time projection. Across six binary and three ternary subsystems, the baseline model achieves a macro-F1 score of 0.951 and 93.98% exact-set match, while physics-informed decoding improves robustness and raises exact-set accuracy to about 96% on dense in-domain grids. The surrogate also generalizes to an unseen ternary section with 99.32% exact-set accuracy and to a quaternary section at 700 °C with 91.78% accuracy. These results demonstrate that attention-based graph learning coupled with thermodynamic constraint enforcement provides an effective and physically consistent surrogate for high-resolution phase mapping and extrapolative alloy screening.

## 1. Introduction
Phase diagrams are foundational to alloy design because they encode equilibrium phase stability as a function of composition and processing conditions, thereby guiding microstructure control and property optimization [1]. However, constructing reliable multicomponent phase diagrams remains expensive, requiring dense composition-temperature sampling and extensive experiments [1]. CALPHAD assessments alleviate this burden by leveraging curated thermodynamic databases and Gibbs-energy minimization, yet dense sweeps over many subsystems or higher-dimensional sections can still be computationally demanding [1,2]. This motivates machine learning (ML) surrogates that can rapidly map composition and temperature to phase information while approximating CALPHAD-level fidelity [3-5]. Open and reproducible CALPHAD workflows have seen tremendous acceleration with tools such as pycalphad, which provides scriptable database parsing and multicomponent multiphase equilibrium calculations [6,7].

These developments naturally motivated machine learning (ML) surrogates that learn phase-diagram behavior from CALPHAD-generated labels and descriptors to reduce evaluation cost [3-5]. A representative early direction framed phase-diagram predictions as classification over phase counts on ternary isothermal sections, showing that descriptor design and ensemble learning can generalize to "held-out" ternaries [3]. A related line of work targeted specific boundaries such as the binary liquidus and miscibility-gap characteristics using gradient boosting and other classical ML models trained on large sets of CALPHAD assessments [4]. In parallel, active-learning phase-diagram construction algorithms (e.g., PDC) and user-facing tools (AIPHAD) emphasized uncertainty-guided sampling and visualization to reduce labeling effort [5,8]. More recently, foundation-model-style approaches have been explored, including fine-tuning and open-source LLMs on CALPHAD-derived question and answer (Q&A) pairs for alloy phase-diagram prediction tasks [9]. Beyond descriptor-based models, graph neural networks (GNNs) have become widely adopted in materials informatics. Crystal-graph neural networks, in particular, learn structure-property relationships directly from atomic connectivity, providing a flexible alternative to hand-crafted descriptors [10].

Subsequent work further generalized this paradigm into a broader graph-network framework applicable to both molecules and crystals, reinforcing GNNs as a standard backbone for representation learning of materials [11]. More recently, the Atomistic Line Graph Neural Network (ALIGNN) showed that explicitly incorporating higher-order interactions (e.g., bond-angle information) via line-graph message passing can improve predictive fidelity, highlighting the benefit of richer relational inductive bias beyond pairwise edges [12] Attention-based variants have also been explored for material prediction: GNNs augmented with global attention (e.g., GATGNN) improved performance by enabling the model to reweight element/atom interactions in a context-dependent manner [13]. Despite this progress, many studies simplified the learning target (e.g., phase count, liquidus temperature, or boundary discovery) rather than predicting the explicit phase set at each composition-temperature point. Predicting the explicit phase assemblage at each state point was formulated as a multi-label learning problem, where each sample was associated with a set of simultaneously valid labels and requires metrics designed for label-set predictions [14]. This distinction matters because microstructure and reliability depend on which phases coexist, not merely how many, and because downstream interpretation (tie-lines/triangles, competition between intermetallic compounds, etc.) requires phase identities [15]. A second practical gap is physical admissibility: purely data-driven multi-label classifiers can output chemically impossible or thermodynamically infeasible phase combinations, especially near boundaries where probabilities are diffuse. Physics-informed learning has been shown to improve stability and correctness in phase-coexistence settings, motivating constraint-aware approaches for phase-diagram surrogates [16].

In this paper, we targeted these two gaps, i.e., explicit phase-set prediction and thermodynamic admissibility in the technologically important quaternary solder system Ag-Bi-Cu-Sn. This system comprises terminal solid solutions of Ag, Cu, Sn, and Bi, a stable liquid phase, and several ordered intermetallic compounds. The stability fields of these intermetallic compounds are narrow and strongly coupled across binary and ternary subsystems, leading to abrupt phase appearances, tightly constrained multiphase equilibria, and sharp transitions with composition and temperature. Such features pose a significant challenge for data-driven models, which must correctly predict discrete phase emergence and coexistence while respecting thermodynamic constraints; unconstrained learning approaches are therefore prone to producing physically inadmissible phase sets in this regime.

We constructed a large equilibrium dataset using CALPHAD calculations executed with pycalphad, leveraging assessed solder thermodynamic resources to obtain phase fractions and derived multi-label phase-presence targets [17]. To represent each state point, we adopted a compact "element graph" with four nodes (Ag, Bi, Cu, Sn) and fully connected edges, reflecting the fact that phase stability is governed by cross-element interactions. For node features, we combined composition (atomic fraction) with Magpie-style elemental attributes, which have demonstrated broad utility as physically meaningful composition descriptors in materials ML [18,19]. We then employed an attention-based GNN backbone using GATv2 layers, benefiting from dynamic attention that has been shown to be more expressive than static GAT attention in capturing context-dependent interactions [20,21]. Most importantly, we formulated the task as multi-label phase-set prediction, because equilibrium CALPHAD solutions often contain multiple coexisting phases at the same state point. For optimization, we adopted modern regularization and scheduling practices (e.g., AdamW and cosine-annealing learning rates) and selected hyperparameters with Optuna to avoid ad-hoc tuning [22-24]. This decision is because random search is often more efficient than grid search for hyperparameter optimization, particularly when only a subset of hyperparameters strongly influences performance [25].

Beyond accuracy, we explicitly enforced thermodynamic admissibility using three lightweight constraints: a Gibbs phase rule-based cap on the number of coexisting phases at fixed temperature and pressure, local smoothness away from boundaries, and pure phase feasibility at composition corners [15]. We implemented these principles in two complementary ways: (i) as optional physics-informed penalties during training (applied one at a time to avoid competing gradients) and (ii) as a deterministic inference-time projection that edits predicted label sets into a feasible set [2]. This separation mirrors a practical observation in physics-informed modeling: training-time regularization can stabilize learning dynamics, while inference-time projection can guarantee feasibility without introducing gradient conflict [16]. Figure 1 summarizes our end-to-end pipeline consisting of: CALPHAD data generation → descriptor/graph construction → GATv2 training → optional physics-aware loss and/or decoding → evaluation on in-domain dense grids and unseen-domain extrapolations.

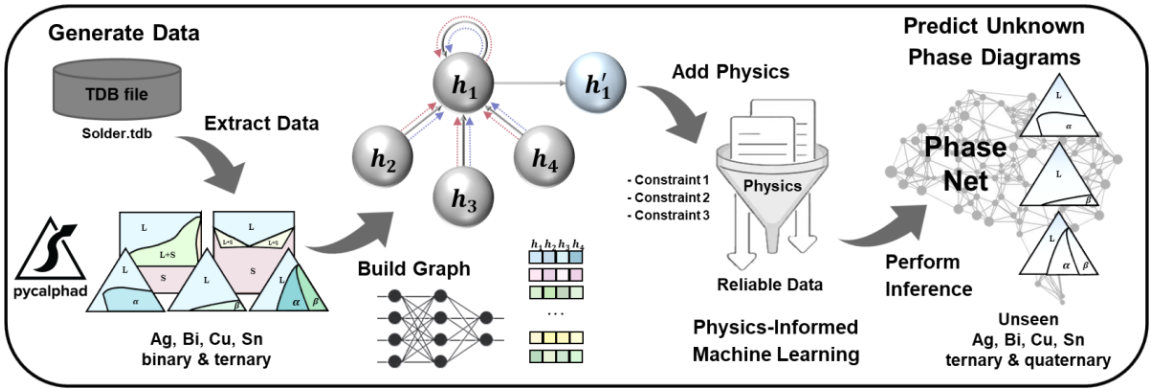

**Figure 1. End-to-end workflow for physics-consistent phase-set prediction.**

Our evaluation strategy reflects both "materials realism" and ML rigor: we assessed performance on trained binary/ternary subsystems, reserved an unseen ternary for extrapolative testing, and further validated with dense grids that stress-tested boundary resolution. In doing so, we focused on providing a surrogate that is not only accurate in aggregate metrics but also reliable in the sense of producing phase maps without frequent unphysical artifacts. Practically, this phase-set formulation provides a direct bridge to downstream metallurgy tasks such as identifying multiphase regions, comparing predicted tie-structures qualitatively, and screening compositions for mitigating the formation of unwanted intermetallics. Finally, because the constraints used here are lightweight and database-agnostic, the same blueprint can be reused across other alloy systems and thermodynamic databases, offering a scalable route toward fast, physics-faithful phase mapping in higher-component spaces [2].

## 2. Methods

This section starts with the generation of an equilibrium dataset for Ag-Bi-Cu-Sn using pycalphad with the NIST solder thermodynamic database [1,6] and explains how phase-presence labels and input descriptors are constructed. We then introduce our element-level graph representation and the GATv2-based multi-label classifier, together with training details and Optuna-based hyperparameter selection. Next, we describe three lightweight thermodynamic constraints and how they are enforced via (i) training-time physics-informed penalties and (ii) inference-time projection. We conclude this section with the evaluation protocol, including seed ensembling, validation-based thresholding, and multi-label performance metrics.

## 2.1 Dataset generation

In this work, we focused on the quaternary Ag-Bi-Cu-Sn alloy system, which is widely used in solder applications, and constructed a dataset comprising approximately 25,000 composition-temperature points. All thermodynamic data was generated with the open-source thermodynamic package pycalphad, which implements CALPHAD (CALculation of PHAse Diagrams) type models [26]. For each sampled composition-temperature point in the Ag-Bi-Cu-Sn system (including the corresponding binary and ternary combinations), we employed pycalphad to perform Gibbs-energy minimization based on the thermodynamic database, thereby obtaining the equilibrium phase assemblage and the associated phase fractions [27]. The CALPHAD-type thermodynamic database was curated by the U.S. National Institute of Standards and Technology (NIST) - the NIST solder alloy CALPHAD Database (Section 2.1.1) [1,6]. We focused on nine relevant phases believed to be important for microstructural evolution in this system: LIQUID, FCC_A1, HCP_A3, BCC_A2, RHOMBO_A7, BCT_A5, EPSILON, Cu-Sn intermetallics, and DO3. For Cu-Sn intermetallic, several distinct CuSn-type compounds exhibited very sparse data coverage in our sampling and were therefore merged into a single aggregated Cu-Sn intermetallic compound to avoid excessively underrepresented labels. All calculations were performed on Dell Precision 3460 workstation with an Intel Core i7-14700 (2.10 GHz) CPU and 32 GB RAM.

### 2.1.1 The pycalphad descriptor

Pycalphad is a widely used open-source Python library implementing the CALPHAD framework, and the calculated phase diagrams closely reproduce reported experimental boundaries. The sampling grid in composition space was constructed such that the four elemental fractions sum to unity, using a step size of 2 at.% for each component. For binary subsystems, temperatures were sampled at intervals of 20 °C over the range relevant to soldering and intermetallic compound formation, with a refined spacing of 10 °C in regions exhibiting frequent phase changes. For the ternary subsystems, data was extracted only on an isothermal section at 700 °C. The resulting phase fractions were used directly as weak labels for multi-label classification, which allows multiple phases to coexist at a given composition-temperature point.

All input variables were subsequently normalized to facilitate stable and efficient training of the machine learning model. For each state point, the presence of individual phases was binarized from pycalphad phase fraction: if the calculated phase fraction exceeded a small threshold ($\varepsilon_{phase} = 10^{-6}$), the corresponding phase was marked as present. The final dataset was randomly split into training, validation, and test sets with a ratio of 80/10/10, while enforcing a minimum number of positive samples per phase in the validation and test splits to avoid extremely under-represented labels.

### 2.1.2 The "Magpie" descriptor

To provide physically meaningful node features for the GNN, we adopted a simplified Magpie-based compositional descriptor [18]. Magpie descriptors capture key elemental properties including atomic number, electronegativity, atomic radius, valence electron counts, and cohesive energy, which are known to influence phase stability and alloy behavior. Combining these intrinsic elemental attributes with the atomic fractions of each constituent, the resulting node features encode both composition and physics-relevant chemical tendencies, enabling the graph network to learn interactions that reflect the underlying thermodynamics of the system. Hence, for each element $e \in \{Ag, Bi, Cu, Sn\}$, we collected eight atomic properties including melting temperature, boiling temperature, density, atomic radius, atomic weight, covalent radius, electronegativity, and first ionization energy, to construct an 8-dimensional property vector. These properties were standardized across the four elements using $z-score$ normalization where each property $\psi$ was transformed as:

$$z = \frac{\psi - \mu}{\sigma}$$

with $\mu$ and $\sigma$ denoting the mean and standard deviation of the property across the four elements. This standardization ensured that each feature had zero mean and unit variance, providing comparable scaling for model training. For each sample, the atomic fraction $x_e$ of each element was concatenated with $\mathbf{z}_e$ to form the node feature $\mathbf{X}_e = [x_e; \mathbf{z}_e] \in \mathbb{R}^9$. The complete list of node and global descriptors is summarized in Table 1.

**Table 1. GNN node and input descriptors for phase-set prediction.**

| Property | Abbreviation | Source |
|---|---|---|
| Atomic fraction of Ag | $x_{Ag}$ | pycalphad |
| Atomic fraction of Bi | $x_{Bi}$ | pycalphad |
| Atomic fraction of Cu | $x_{Cu}$ | pycalphad |
| Atomic fraction of Sn | $x_{Sn}$ | pycalphad |
| Melting temperature (K) | $T_{Melt}$ | Magpie |
| Boiling temperature (K) | $T_b$ | Magpie |
| Density ($g/cm^3$) | $\rho$ | Magpie |
| Atomic radius (pm) | $r_{atom}$ | Magpie |
| Atomic weight ($g/mol$) | $M$ | Magpie |
| Covalent radius (pm) | $r_{cov}$ | Magpie |
| Electronegativity (Pauling) | $\chi$ | Magpie |
| First ionization energy (eV) | $IE_1$ | Magpie |
| Global temperature | $T$ | Input state variable |

## 2.2 Graph neural network (GNN) architecture

Graph-based materials models showed that learnable message passing can effectively capture interaction patterns underlying material responses [10,11]. Moreover, extensions such as ALIGNN demonstrated that explicitly modeling higher-order interactions (e.g., angle information) could further improve predictive fidelity, highlighting the benefit of expressive interaction modeling in material learning [12]. Although these models were often formulated on atomistic or crystal graphs, the same core idea, which is learning interaction structure from data, also applies to composition driven phase stability, where cross-element interactions govern equilibria. Graph attention models augmented with global attention (e.g., GATGNN) had demonstrated improved performance in inorganic material property prediction, supporting the use of attention mechanisms to capture both local interactions and global contributions in graph representations [13] Attention mechanisms provided a flexible way to compute the context-dependent importance weights, enabling models to selectively emphasize informative interactions in a data-driven manner [28].

In this work, we leveraged GATv2 on a fully connected element graph to learn composition-dependent interaction weights among Ag, Bi, Cu, and Sn [21]. First, each composition-temperature state was encoded as a four-node fully connected element graph with Magpie-based node features. Next, the graph was processed by stacked GATv2 layers to obtain node embeddings, which were pooled into a graph-level representation and combined with temperature before a multi-layer perceptron (MLP) provided multi-label phase-presence probability outputs.

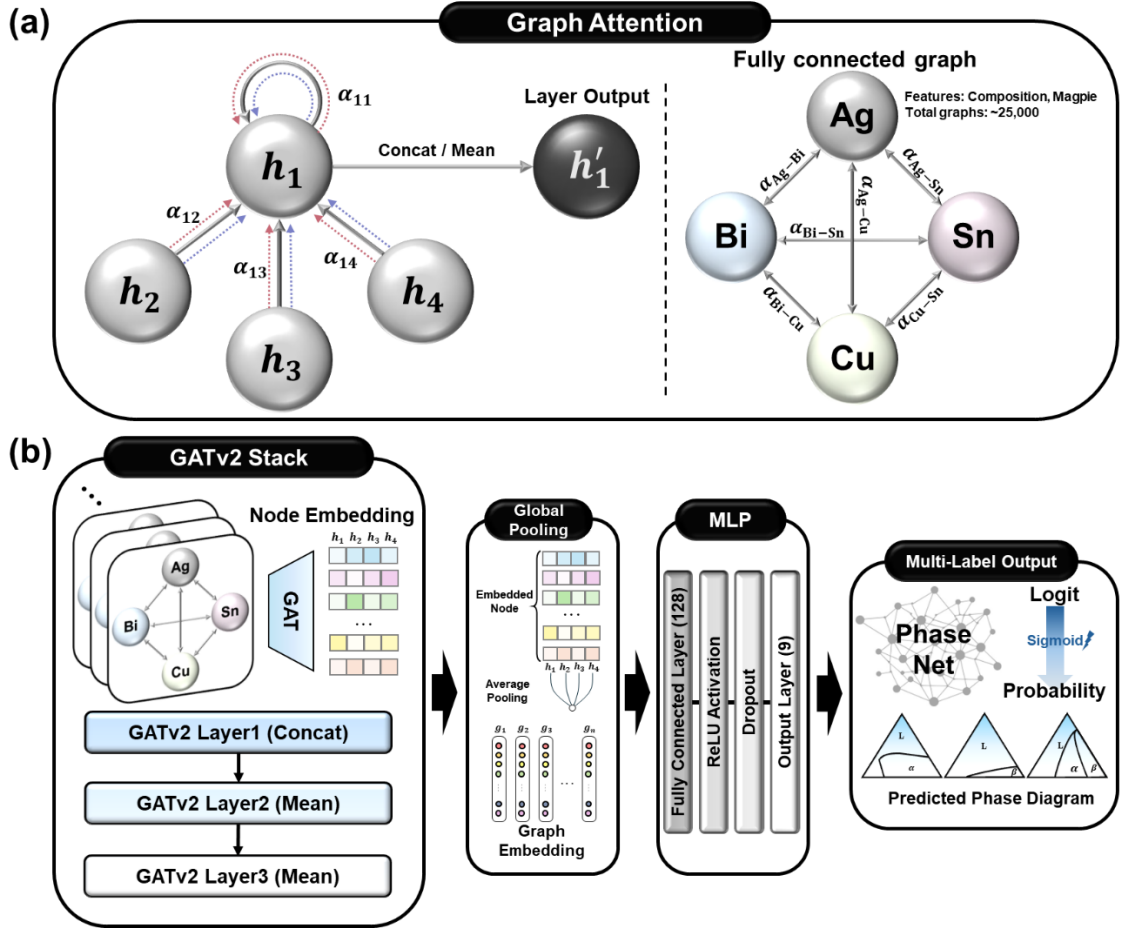

**Figure 2. GATv2-based architecture for multi-label phase-set prediction. (a) Graph attention layer: Fully connected element graph (Ag/Bi/Cu/Sn); node features = mole fractions and eight-dimensional Magpie; GATv2 computes attention and updates embeddings. (b) End-to-end backbone: GATv2 encodes element; pooled features with temperature drive an MLP to phase probabilities.**

### 2.2.1 Element-graph construction and GATv2 attention layer

Figure 2(a) provides a guide to how a single composition-temperature sample was converted into an element graph and processed by one GATv2 attention layer. The left panel of Figure 2(a) shows the message passing on the 4-node graph; while the right panel summarizes the elemental constituents (Ag, Bi, Cu, Sn) that instantiate the four nodes for this alloy system. With 25,000 composition-temperature samples, we constructed 25,000 graphs. For each sample, we built a graph $G = (V, E)$, where $V$ is the set of nodes and $E$ is the set of edges. Specifically, we used a fully connected directed graph with $|V| = 4$ nodes corresponding to Ag, Bi, Cu, and Sn. The node-feature matrix is denoted by:

$$\mathbf{X} = \begin{bmatrix} \mathbf{X}_{Ag}^T \\ \mathbf{X}_{Bi}^T \\ \mathbf{X}_{Cu}^T \\ \mathbf{X}_{Sn}^T \end{bmatrix} \in \mathbb{R}^{4 \times d_{in}} \quad \text{where } d_{in} = 9,$$

where $\mathbf{X}_e = [x_e; \mathbf{z}_e]$ concatenates the atomic fraction $x_e$ and the eight-dimensional Magpie $\mathbf{z}_e$ of element $e$ (Section 2.1.2). Let $\mathbf{H}^{(0)} = \mathbf{X}$ denote the initial node-feature matrix input to the mode, and $\mathbf{H}^{(l)} \in \mathbb{R}^{4 \times d_l}$ be the node-embedding matrix at layer $l$.

At each GATv2 layer $l$, the embedding of node $i$ was updated by attention-weighted aggregation of neighbor information [29]:

$$\mathbf{h}_i^{(l+1)} = \sigma \left( \sum_{j \in \mathcal{N}_i} \alpha_{ij}^{(l)} \mathbf{W}^{(l)} \mathbf{h}_j^{(l)} \right),$$

where $\mathcal{N}_i$ denotes the set of neighbors of node $i$, $\mathbf{h}_i^{(l)}$ is the embedding vector of node $i$ at layer $l$, $\mathbf{W}^{(l)}$ is a learnable weight matrix, $\alpha_{ij}^{(l)}$ are the attention coefficients, and $\sigma(\cdot)$ denotes the non-linear activation function. In GATv2, the attention coefficients $\alpha_{ij}^{(l)}$ were obtained in two steps. First, an unnormalized attention score $e_{ij}^{(l)}$ was computed for each edge $(i,j)$ using the linearly projected node features:

$$e_{ij}^{(l)} = \text{LeakyReLU} \left( \boldsymbol{a}^\mathrm{T} \left[ \mathbf{W}^{(l)} \mathbf{h}_i^{(l)} \| \mathbf{W}^{(l)} \mathbf{h}_j^{(l)} \right] \right),$$

where $\boldsymbol{a}$ is the learnable attention vector, $\|$ denotes concatenation, and LeakyReLU is the non-linear activation function. Second, these unnormalized scores were normalized across all neighbors $j$ of node $i$ using the softmax function to obtain the attention coefficients:

$$\alpha_{ij}^{(l)} = \frac{exp\,(e_{ij}^{(l)})}{\sum_{k \in \mathcal{N}_i} exp\,(e_{ik}^{(l)})}.$$

### 2.2.2 End-to-end backbone, pooling, and training objective

Figure 2(b) summarizes the end-to-end prediction pipeline. The 4-node element graph was encoded by three stacked GATv2 layers (four heads each). We then obtained a graph-level embedding by global mean pooling over the four nodes:

$$\boldsymbol{g} = \frac{1}{|V|} \sum_{i \in V} \mathbf{h}_i^{(3)} \in \mathbb{R}^{160}.$$

We concatenated this embedding with the temperature feature $T \in \mathbb{R}^1$ to form $[\boldsymbol{g}; T] \in \mathbb{R}^{161}$, which was passed to an MLP. The graph-level embedding dimension was set to 160. Rather than an arbitrary choice, this dimension corresponds to the optimal hidden size identified through our Optuna hyperparameter optimization (detailed in Section 2.2.3), providing the best balance between model expressivity and generalization stability. The MLP consisted of a fully connected layer, ReLU activation, dropout, and an output layer that produced nine phase-presence logits. Dropout was applied to mitigate overfitting by preventing co-adaptation of hidden units [30].

We trained the model in a multi-label setting. During training, we applied a sigmoid function to the logits and optimized a class-balanced focal loss. At evaluation time, per-class decision thresholds were tuned on the validation set to maximize the Macro-F1 score. For optimization, we used AdamW [21,31] with cosine-annealing learning-rate scheduling. To ensure reproducibility, we ran 10 independent training runs with different random seeds and reported the Macro-F1 score across seeds.

### 2.2.3 Optuna automatic hyperparameter tuning

To avoid manual trial-and-error in setting model and training hyperparameters, we employed Optuna, an automatic hyperparameter optimization framework [24]. We formulated hyperparameter tuning as a black-box optimization problem. Bayesian optimization provides a principled framework for black-box hyperparameter tuning by modeling validation performance and selecting promising configurations sequentially [32]. Let $\lambda_{Optuna}$ denote a vector of hyperparameters (e.g., learning rate, weight decay, hidden dimension, dropout ratio, and batch size). For a given $\lambda_{Optuna}$, we defined the objective function $J(\lambda_{Optuna})$ as the best Macro-F1 score achieved on the validation set across all training epochs. Optuna searched for the optimal hyperparameter configuration $\lambda^*$ that maximizes this objective:

$$\lambda^* = \underset{\lambda_{Optuna}}{\mathrm{argmax}}\, J(\lambda_{Optuna}).$$

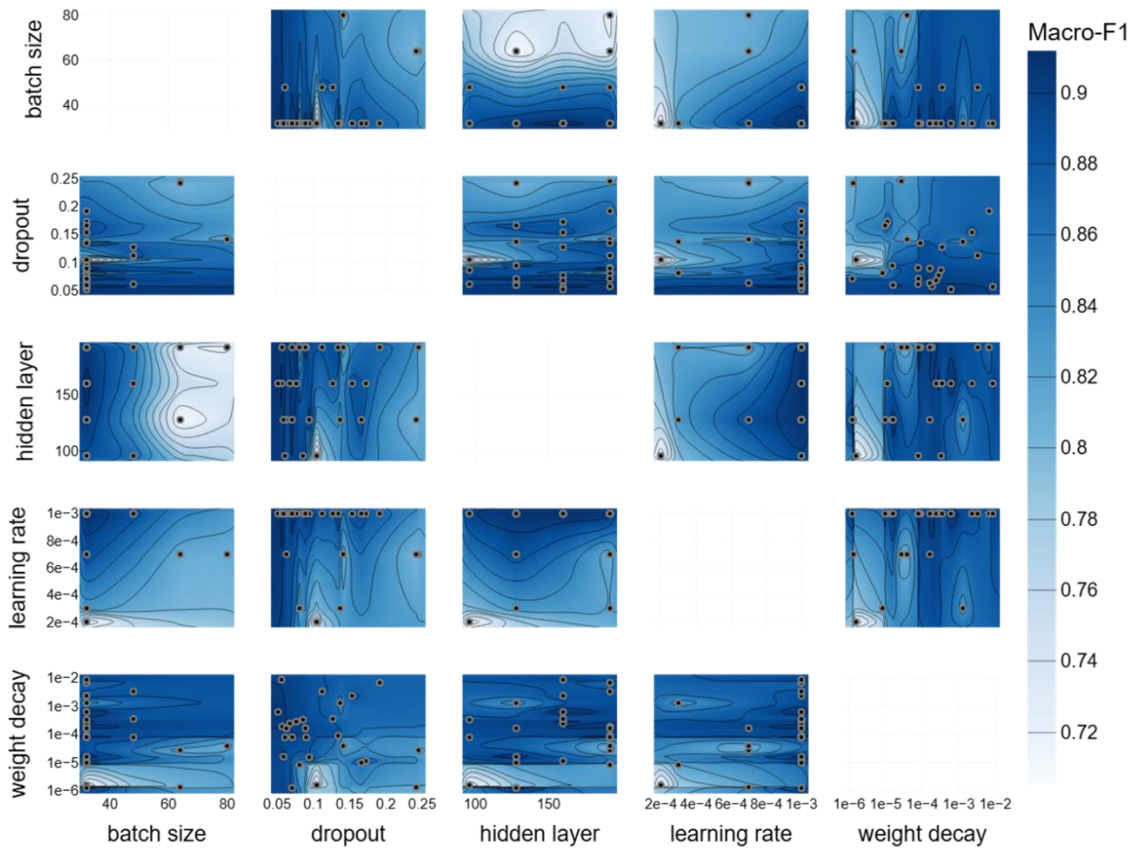

**Figure 3. Pairwise hyperparameter interaction contour of the Macro-F1 score from Optuna.**

For each iteration, Optuna sampled a candidate set of hyperparameters, trained the GNN model on the training split, and reported the best validation Macro-F1 score obtained during training as the objective value. The search covered typical ranges for the learning rate, weight decay, hidden dimension of the GATv2 layers, dropout ratio, and parameters of the loss function, while keeping the overall architecture in Figure 2. Each candidate model was trained for up to 100 epochs with early stopping based on the validation Macro-F1 score and a cosine-annealing learning rate schedule. Early stopping based on validation performance was used to avoid overfitting and to select a stopping point that improves generalization [33]. The best set of parameters selected by Optuna was thereafter used in this work, namely a hidden dimension of 160, batch size of 32, learning rate $1.0 \times 10^{-3}$, dropout 0.05, weight decay $6.0 \times 10^{-4}$ with the class-balanced focal loss.

Figure 3 summarizes the pairwise hyperparameter interactions evaluated during the Optuna search. These contour plots map the objective value (Macro-F1 score) across varying combinations of critical hyperparameters, such as learning rate, hidden dimension, dropout ratio, and weight decay. The contour gradients effectively highlight the regions within the search space that yield the highest predictive performance, revealing the optimal operating boundaries for the GNN model. Furthermore, Figure 4 depicts the optimization history, tracing the validation Macro-F1 score across successive trials. The trajectory illustrates how the Bayesian optimization algorithm efficiently explores the search space during early iterations before exploiting promising regions. Eventually, the objective value converges and stabilizes, identifying the optimal hyperparameter configuration used in our final evaluations.

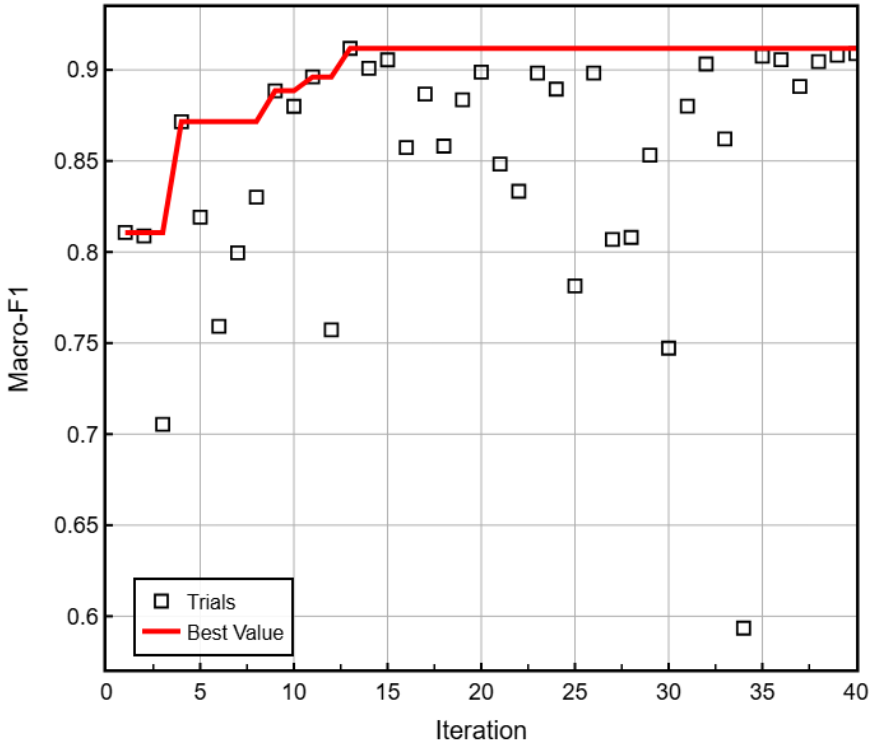

**Figure 4. Optuna optimization history depicting the evolution of Macro-F1 score with iterations.**

### 2.3 Implementation of physics constraints
In this work, we implemented multiple physics constraints to promote physically admissible phase-set predictions and to reduce spurious multiple-phase activations. These constraints were incorporated both as physics-informed loss penalties during training and as a deterministic, constraint-aware decoding step at inference, thereby improving physical fidelity and predictive robustness.

**Gibbs phase rule (GPR):** At thermodynamic equilibrium, the number of coexisting phases is constrained by the Gibbs phase rule $F = C - P + 2$, where $F$ is the thermodynamic degrees of freedom, $C$ is the number of elements, and $P$ is the number of phases [34]. In our targeted alloy system, the elements correspond directly to the elemental constituents (Ag, Bi, Cu, Sn). Under fixed temperature and pressure, this reduces to $F = C - P \geq 0$, implying that the number of coexisting phases cannot exceed the number of chemically present elements. Thus, binaries allow at most two phases and ternaries at most three. In our representation, we estimated the number of present elements for $n$-th sample, $C_n$, as

$$C_n = \sum_e 1\left(x_e^{(n)} > \varepsilon_{element}\right), \qquad \varepsilon_{element} = 10^{-6},$$

where $n$ is the index of the data sample, $e$ denotes the elements, $x_e^{(n)}$ is the atomic fraction of element $e$ in sample $n$, and $\varepsilon_{element}$ is a small threshold to ignore numerical noise. Practically, this constraint suppressed spurious multi-phase activation (e.g., three phases in binary) and aligned the model with equilibrium CALPHAD outputs.

**Local smoothness**: Away from phase boundaries, equilibrium phases and fractions vary smoothly with composition and temperature because Gibbs free energy surfaces are analytic and chemical potentials change continuously [19]. From a learning-theoretic perspective, this corresponds to a smoothness prior on the predicted label function over the composition-temperature manifold, closely related to manifold regularization where neighborhood structure is used to penalize rapid function variation and stabilize predictions [35]. We operationalized this by a Gaussian neighborhood weight ($\omega_{nm}$) between samples $n$ and $m$. Let $\mathbf{x}_n, \mathbf{x}_m \in \mathbb{R}^4$ denote the composition vectors (mole fractions of Ag/Bi/Cu/Sn) for sample $n$ and $m$, respectively, and let $T_n, T_m \in \mathbb{R}$ denote its temperatures (normalized to [0,1] over the dataset range). We defined

$$\omega_{nm} = \exp\left(-\frac{\|\mathbf{x}_n - \mathbf{x}_m\|_2^2}{\sigma_x^2} - \frac{(T_n - T_m)^2}{\sigma_T^2}\right),$$

with $\sigma_x = \sigma_T = 0.05$ (5% normalized distance). To ensure computational efficiency and prevent excessive blurring across sharp first-order phase boundaries, we strictly limited this neighborhood computation to the eight nearest neighbors for each state point. This localized scale effectively smoothed out isolated noise artifacts while perfectly preserving the abrupt transitions characteristic of the multi-phase equilibria.

**Pure phase feasibility**: At compositions where exactly one element is present

$$P = \{n \in B \mid \sum_e 1\left(x_e^{(n)} > \varepsilon_{element}\right) = 1\}, \quad \varepsilon_{element} = 10^{-6},$$

the system reduces to a unary phase diagram: at any fixed temperature a single phase is thermodynamically stable (allowing temperature-driven allotropes across $T$, but not simultaneously at the same $T$). Here, $B$ denotes the set of all evaluated composition points in the dataset. As defined previously, $x_e(n)$ denotes the fraction of element $e$ in sample $n$. And the parameter $\varepsilon_{element}$ was set to a very small threshold to numerically distinguish elements that were effectively present from those that were absent. The condition $\sum_e 1(x_e^{(n)} > \varepsilon_{element}) = 1$ therefore ensured that only one element had a non-negligible fraction in the composition, while all other elements were essentially zero. As a result, the set $P$ defined by this condition was restricted to the set of pure elements, or unary systems, where each composition contains only a single element in significant proportion.

Thus, predictions at corner compositions were represented as one-hot vectors. In terms of implementation, we parsed phase names to restrict the admissible phases strictly to those compatible with the present pure element; generic lattices (e.g., LIQUID, FCC_A1) were treated consistently with the CALPHAD database. This prevented corner activation of intermetallic that required absent elements. In physics-informed losses, each principle was instantiated once as a penalty term added to the supervised loss, trained one at a time to avoid competing gradients and to isolate effects. The same principles were enforced as physics-informed decoding applied post-threshold: we first pruned impossible corner labels (pure phase feasibility), then applied strictly local probability smoothing, and finally imposed the cardinality cap (Gibbs phase rule). This sequence was safe to combine because it edited outputs without altering learned decisions.

The aforementioned physical constraints were incorporated through both physics-informed loss functions and physics-informed decoding, as illustrated in Figure 5. In the physics-informed loss formulation, the model was penalized for predictions that violated known physical laws or constraints, thereby guiding the learning process toward physically consistent solutions. In physics-informed decoding, physical constraints were enforced directly at the output stage of the network. The unconstrained model predictions were transformed through constraint-aware decoding operations, such as normalization, bounding, or projection onto a physically admissible space, ensuring that the final outputs strictly satisfy the required physical conditions. This dual strategy combined soft constraint enforcement during training with hard constraint satisfaction at inference, improving both physical fidelity and predictive robustness.

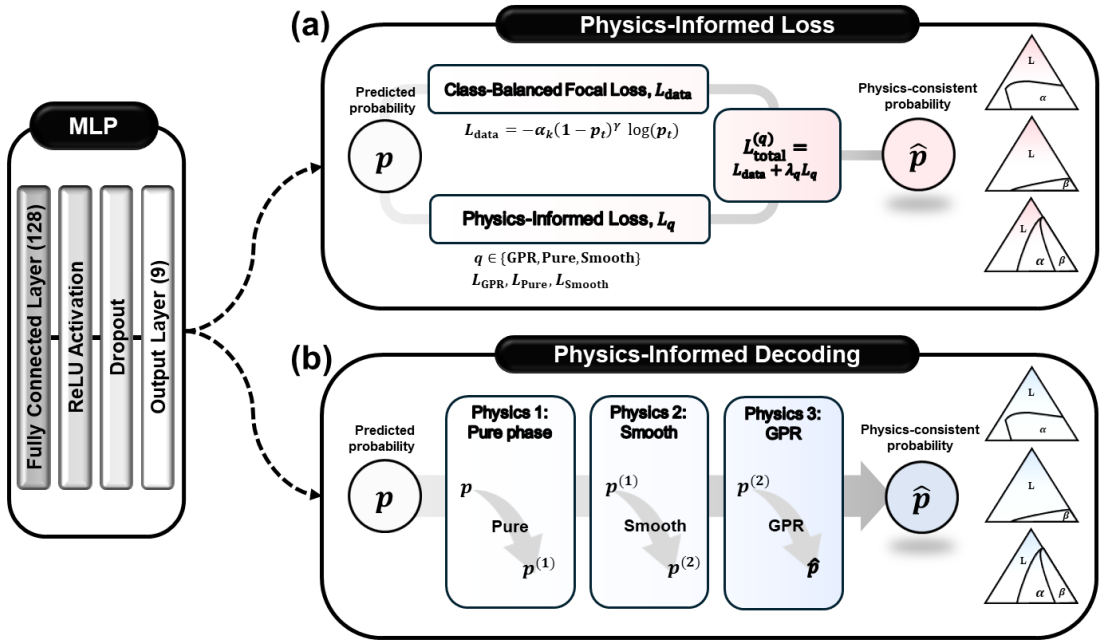

**Figure 5. Schematic of physics-informed training and decoding. (a) The supervised loss was combined with a physics-informed penalty to form the total training objective. (b) Physics-informed decoding refined the raw predictions so that they obey physical constraints on admissibility of phase sets. In this work, the Ag-Bi-Cu-Sn phase diagram was governed by three thermodynamic principles. Beyond improving accuracy, these constraints shrank the hypothesis space of admissible labeling, thereby improving data efficiency and stabilizing training and inference.**

### 2.3.1 Physics-informed loss

Let $n \in \{1, \ldots, N\}$ be the index of a training sample, where $N$ is the total number of training samples, and $k \in \{1, \ldots, K\}$ be the index of a specific phase class, where $K = 9$ is the total number of target phases. For the $n$-th sample, let $\mathbf{y}_n \in \{0,1\}^K$ denote the ground-truth phase presence vector.

The GNN output is a logit vector $\mathbf{o}_n \in \mathbb{R}^K$. After applying the sigmoid transformation, we obtained the predicted probability vector $\mathbf{p}_n = \sigma(\mathbf{o}_n) \in [0,1]^K$, where the $k$-th element, $p_{n,k}$, represents the predicted probability that phase $k$ is present in sample $n$. We minimized a class-balanced focal loss:

$$L_{\text{data}} = -\frac{1}{N} \sum_{n=1}^{N} \sum_{k=1}^{K} \beta_k (1 - p'_{n,k})^\gamma \log(p'_{n,k})$$

where $p'_{n,k} = p_{n,k}$ if $y_{n,k} = 1$, and $p'_{n,k} = 1 - p_{n,k}$ if $y_{n,k} = 0$. For clarity, the simplified per-sample loss is shown in Figure 5(a). Here, $\beta_k$ is a class-specific weighting factor to handle class imbalance, and $\gamma$ is the focusing parameter that down weights well-classified examples to concentrate the model on difficult instances. This formulation allowed our model to handle class imbalance while concentrating on learning difficult examples. Each physics-informed constraint was instantiated as an additional penalty $L_q$, yielding the total loss

$$L_{\text{total}}^{(q)} = L_{\text{data}} + \lambda_q L_q, \qquad q \in \{\text{GPR, smooth, pure}\},$$

where $\lambda_q$ is a loss weighting hyperparameter that controls the importance of the physics constraint relative to the data loss. To isolate the effects of individual constraints, we trained with one physics-informed loss at a time, ensuring that competing gradients did not interfere with the learning of each principle. The overall training and inference strategies are shown in Figure 5(a).

**Gibbs phase rule loss ($L_{GPR}$):** Let $p_{n,k} \in [0,1]$ denote the predicted presence probability of phase $k$ for sample $n$ and $C_n$ is the estimated number of chemically present elements (Section 2.3, Gibbs phase rule). We defined $S_n = \sum_{k=1}^{K} p_{n,k}$ as a soft surrogate for the predicted number of coexisting phases. The Gibbs phase rule penalty

$$L_{GPR} = \frac{1}{|\mathcal{B}|} \sum_{n \in \mathcal{B}} [\max(0, S_n - C_n)]^{\gamma_{GPR}}$$

penalized samples whose total phase activation exceeded the admissible cap $C_n$. Here, $\gamma_{GPR} \geq 1$ controlled how strongly large violations were emphasized, and $\mathcal{B}$ denotes a training batch with $|\mathcal{B}|$ being the number of samples in the batch.

**Local smoothness loss ($L_{smooth}$):** Let $\omega_{nm}$ be the Gaussian neighborhood weight defined in Section 2.3, which is large only for nearby composition-temperature points. The smoothness penalty

$$L_{smooth} = \frac{\sum_{n,m} \omega_{nm} \|\mathbf{p}_n - \mathbf{p}_m\|_2^2}{\sum_{n,m} \omega_{nm} + \delta}$$

encouraged neighboring samples to have similar predicted probability vectors $\mathbf{p}_n \in [0,1]^K$. The normalization by $\sum \omega_{nm}$ made the term insensitive to the number of neighbor pairs, and $\delta$ prevent numerical instability when weights were small.

**Pure phase loss ($L_{Pure}$):** Let $P$ denote the set of "pure-corner" samples (Section 2.3). For $n \in P$, we defined $S_n = \sum_{k=1}^{K} p_{n,k}$ and $M_n = \max_k p_{n,k}$. The penalty

$$L_{Pure} = \frac{1}{|P|} \sum_{n \in P} [\max(0, S_n - M_n)]^{\gamma_{pure}},$$

encouraged one-hot predictions at pure phase by forcing the total activation $S_n$ to match the largest entry $M_n$, thereby discouraging simultaneous activation of multiple phases at unary corners. Here, $\gamma_{pure} \geq 1$ controlled how strongly these violations were penalized, and $|P|$ denotes the number of pure-corner samples present in the current batch.

### 2.3.2 Physics-informed decoding

Physics-informed decoding or projection is a post-training filter that converts real-valued scores into physically admissible label sets without retraining. Deterministic post-hoc projection is also well aligned with a broader ML trend of enforcing constraints via optimization-based operators (e.g., casting constraint satisfaction as an explicit optimization layer), which motivates using principled projection mechanisms to guarantee feasibility beyond what penalty terms can ensure [36]. Unlike adding losses (which can yield competing gradients and dilute supervision), projection is deterministic and safe to combine (hybrid) in a single pass. In our formulation, after training, a fixed sequence of post-processing operators was applied to the model outputs in order to explicitly enforce the three physical constraints.

The first operator enforced pure phase feasibility by suppressing phase predictions that were physically impossible at composition endpoints. This step removed infeasible phase labels at pure or near-pure phases, ensuring consistency with known boundary conditions of the phase diagram. Applying this constraint prevented physically invalid activations from propagating into subsequent operations. Next, local smoothness was imposed by attenuating low-confidence, spatially isolated activations using a Gaussian neighborhood filter. This operation promoted continuity in composition-temperature space by reinforcing phase predictions that were supported by nearby points while suppressing spurious, noise-driven predictions. As a result, phase regions became more coherent and physically plausible.
Finally, a Gibbs cap was enforced via a dynamic cardinality projection that limited the number of simultaneously active phases at each state point. This operator reflected the Gibbs phase rule by imposing a localized constraint on the predicted phase set, ensuring that the number of coexisting phases at equilibrium does not exceed the estimated number of present components ($C_n$).

The ordering of these operators from pure component feasibility → local smoothness → Gibbs cap was critical. Impossible phase labels were first eliminated, then the remaining predictions were stabilized through neighborhood aggregation, and only afterward was the global cardinality constraint applied to a clean and physically admissible candidate set. This structured decoding pipeline improved physical consistency while preserving meaningful model confidence. The overall training and inference strategies are shown in Figure 5(b).

### 2.4 Evaluation protocol

Unless otherwise noted, all metrics were computed on the held-out test split. For each input, we formed an ensemble by averaging per-phase probabilities across seeds (Seed ensembling); per-class decision threshold was tuned on the validation split via grid search to maximize the Macro-F1 score. Ensembling multiple independently trained models was a simple and effective approach to improve predictive robustness and stabilize uncertainty-related variation in neural predictions [37]. In Section 3, we provide (i) per-phase F1s, (ii) overall Macro-F1, (iii) exact-set accuracy for binary (A-B) and ternary (A-B-C) subsets. All metrics are tabulated for each system and summarized as the mean across all A-B pairs and the mean across all A-B-C triplets with non-zero overlap.

**Macro-F1:** In our multi-label setting, each phase $k$ was evaluated as a one-vs-rest binary task. In binary classification, $TP_k$ counted points where phase $k$ was present in both truth and prediction, $FP_k$ when predicted but absent, and $FN_k$ when true but not predicted. The per-class score

$$F1_k = \frac{2TP_k}{2TP_k + FP_k + FN_k}$$

is the harmonic mean of precision and recall (true negatives do not enter), which was appropriate in our analysis when most phases were absent at a given point. Macro-F1 is the unweighted average across phases, giving equal importance to rare and frequent phases, thus mitigating class-imbalance bias. This score is given by:

$$Macro - F1 = \frac{1}{K}\sum_{k=1}^{K} F1_k,$$

where $K$ is the total number of target phases. The calculation was performed by applying a validation-based threshold to the seed average probability and then binarizing it.

**Accuracy (exact-set match):** This measure was adopted to characterize the fraction of samples whose entire multi-label set exactly matched the ground truth. At each evaluated point, predicted and true labels were converted to a sorted phase set string (e.g., "LIQUID+FCC_A1", or "NONE" if empty). The performance was evaluated using the mismatch count ($N_{mismatch}$) and the associated exact-set accuracy. For all our analysis, we calculated the count (and rate) per binary (A-B) and ternary (A-B-C) systems, and in aggregate as follows:

$$\text{Subset Acc} = \frac{1}{N}\sum_{n=1}^{N} 1\{\hat{\mathbf{y}}_n = \mathbf{y}_n\},$$

where $N$ is the total number of evaluated samples in the specific subset, $n$ is the sample index, $\mathbf{y}_n$ and $\hat{\mathbf{y}}_n$ are the ground-truth and predicted phase sets for sample $n$, respectively. Alternatively, this exact-set match rate can be directly expressed in terms of the mismatch count:

$$\text{Accuracy} = 1 - \frac{N_{mismatch}}{N}.$$

## 3. Results and discussion

This section reports (i) overall multi-label phase-set prediction performance across in-domain and extrapolative settings, (ii) the impact of physics-informed losses and decoding on physical admissibility and accuracy, and (iii) qualitative comparisons on dense composition-temperature grids. Before evaluating the physics constraints, we first established an optimal baseline model using Bayesian hyperparameter tuning. As detailed in the methodology, the pairwise hyperparameter interaction contours (Figure 3) and the overall optimization history (Figure 4) demonstrate that the Optuna framework efficiently explored the search space and converged to a stable, high-performing configuration. Through securing this optimal baseline architecture, which was characterized by converged objective values and well-defined hyperparameter boundaries, we ensured that any further performance gains and variance reductions discussed in the following subsections are strictly attributable to the physics-informed strategies rather than sub-optimal baseline tuning.

### 3.1 Comparison of baseline GNN performance with physics-based variants

Table 2 reports Macro-F1 scores for the baseline GNN and its physics-informed variants across binary and ternary subsystems. Overall, all models achieved high predictive performance, indicating that our graph-based representation effectively captured phase stability relationships across composition-temperature space. However, consistent improvements were observed when physical constraints were incorporated, particularly through physics-informed decoding. The baseline GNN achieved an overall Macro-F1 of (0.9513 ± 0.0141), with strong performance across all binary systems (Macro-F1 >0.95) and near-perfect accuracy for ternary systems, especially Ag-Bi-Cu (0.9979). The relatively larger standard deviation might indicate sensitivity to training variability in the unconstrained model.

Incorporating a physics-informed loss improved the overall Macro-F1 score to (0.9577 ± 0.0049), while substantially reducing variance across runs. This indicates that penalizing physically inconsistent predictions during training stabilized learning and enhanced generalization. Notable gains were observed in systems such as Bi-Sn and Cu-Sn, suggesting that explicit physical regularization is particularly beneficial in regions where phase boundaries are more complex or less densely sampled. The best overall performance was achieved with physics-informed decoding, which attains an overall Macro-F1 of (0.9623 ± 0.0035). Improvements were especially pronounced in the ternary subsystems Ag-Cu-Sn and Bi-Cu-Sn, where enforcing pure phase feasibility, local smoothness, and the Gibbs phase constraint at inference time effectively suppressed spurious phase activations. The consistently low variance further demonstrated that hard constraint enforcement during decoding resulted in more robust and physically consistent predictions than training-time regularization alone.

Comparing the two physics-informed approaches, decoding-based constraint enforcement outperformed loss-based regularization, despite both embedding the same physical knowledge. This suggests that directly projecting predictions onto a physically admissible space was more effective than softly penalizing violations during optimization, particularly for enforcing global constraints such as phase cardinality. Overall, these results highlight the complementary roles of physics-informed learning and decoding, while demonstrating that post hoc physics-informed decoding provides the largest and most reliable performance gains across both binary and ternary alloy systems.

**Table 2. Overall and per-system Macro-F1 scores for multi-label phase-set prediction across binary and ternary systems.**

| | Overall Macro-F1 | Binary | | | | | | Ternary | | |
|---|---|---|---|---|---|---|---|---|---|---|
| | | Ag-Bi | Ag-Cu | Ag-Sn | Bi-Cu | Bi-Sn | Cu-Sn | Ag-Bi-Cu | Ag-Cu-Sn | Bi-Cu-Sn |
| **GNN** | 0.9513 ±0.0141 | 0.9862 | 0.9713 | 0.9658 | 0.9906 | 0.9620 | 0.9570 | 0.9979 | 0.9622 | 0.9674 |
| **GNN + Physics-Informed Loss** | 0.9577 ±0.0049 | 0.9894 | 0.9632 | 0.9684 | 0.9962 | 0.9741 | 0.9681 | 0.9982 | 0.9629 | 0.9685 |
| **GNN + Physics-Informed Decoding** | 0.9623 ±0.0035 | 0.9874 | 0.9722 | 0.9713 | 0.9951 | 0.9752 | 0.9667 | 0.9981 | 0.9713 | 0.9701 |

Performance was consistently high for systems characterized by well-defined phase boundaries, while modest reductions were observed in systems with more complex boundary structures. To complement the Macro-F1 results, Figure 6 presents subset accuracy metrics aggregated across systems and further disaggregated by individual subsystems. These metrics corroborate the Macro-F1 trends, demonstrating that physics-informed decoding consistently yielded the highest exact-set match accuracy and lowest variance across varying levels of thermodynamic complexity.

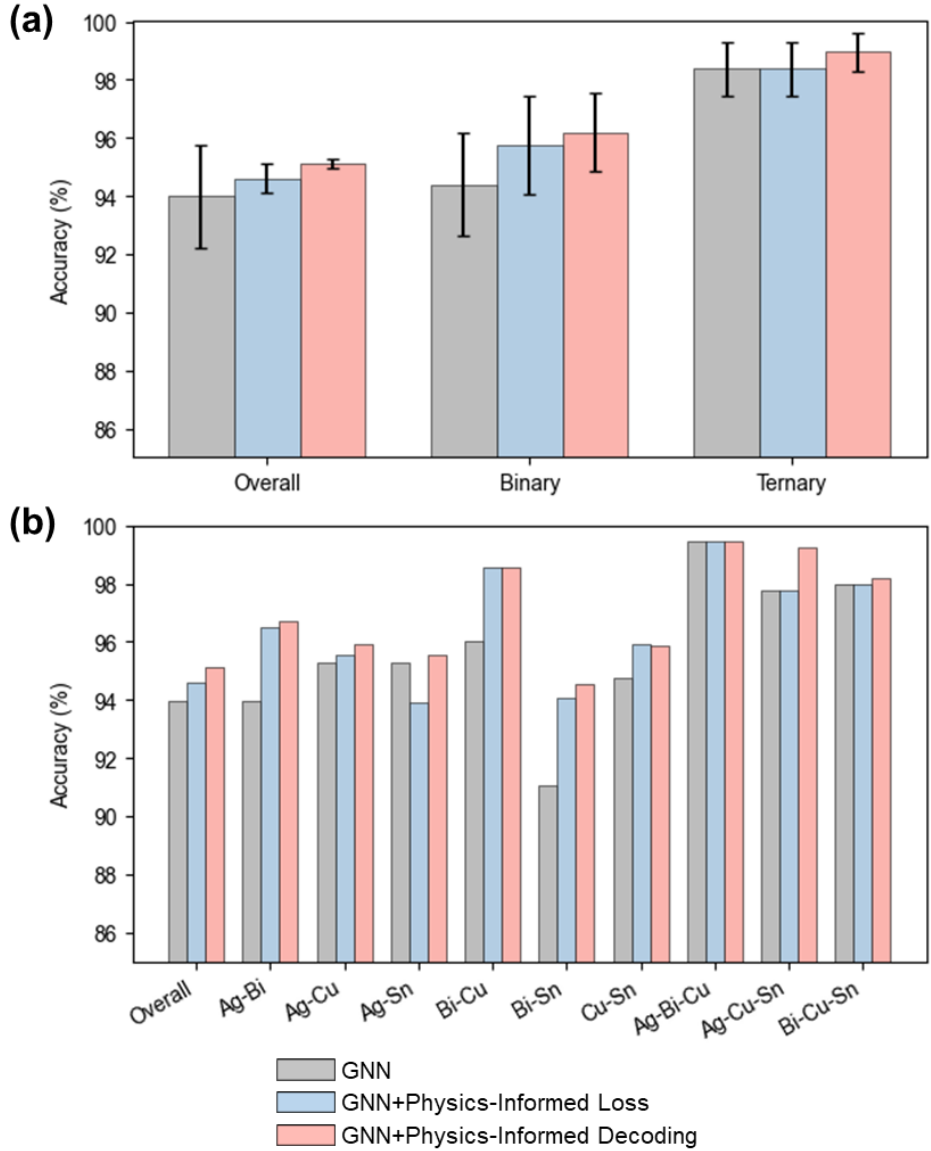

**Figure 6.** Accuracy of phase-set prediction with and without physics constraints. (a) A mean subset accuracy over all binary and ternary systems; error bars indicate the standard deviation over 10 runs. (b) per-system subset accuracy for individual binary and ternary solder alloys.

## 3.2 Evaluation of physics-informed loss and physics-informed decoding

We conducted a controlled ablation study to evaluate the impact of three physics-informed loss terms (Gibbs phase rule, local smoothness, and pure phase feasibility) on model performance. For each loss, we systematically varied the corresponding loss weight $\lambda_q \in \{0.05, 0.10, \ldots, 0.45\}$, where $q \in \{GPR, smooth, pure\}$, while holding all other hyperparameters constant. To ensure a fair comparison and avoid confounding effects from early stopping, each configuration was trained for exactly 20 epochs and evaluated on the validation split using the overall Macro-F1 score as the performance metric. Among the candidates, the Gibbs phase rule loss consistently achieved the highest Macro-F1, with a well-defined optimum at $\lambda_{GPR} = 0.15$. Consequently, for all subsequent analyses, we adopt the Gibbs phase rule with $\lambda_{GPR} = 0.15$ configuration as the representative physics-informed-loss model in head-to-head comparisons with both the unconstrained GNN and the physics-informed decoding approach.

We deliberately avoided combining multiple physics-informed loss terms during training. In practice, applying concurrent constraints could generate competing gradient directions, which might dilute the supervised signal and destabilize optimization, ultimately reducing predictive accuracy. Focusing on a single physics-informed loss mitigated this issue and yielded modest but consistent improvements across metrics: the overall Macro-F1 increased to 0.958 and subset accuracy reached 94.60%. Notably, performance gains were most pronounced in boundary-dense subsystems such as Bi-Sn and Cu-Sn, where physical constraints helped resolve ambiguous predictions. Conversely, predictions in near-ceiling systems such as Ag-Bi-Cu remained largely unaffected, reflecting the already high baseline accuracy in these simpler regions.

The results of the loss-weight sweep are summarized in Table 3. As shown, the Gibbs phase rule loss outperformed both the local smoothness and pure phase penalties across nearly all $\lambda_q$ values and reached a peak at $\lambda_{GPR} = 0.15$, reinforcing its selection for all subsequent physics-informed comparisons. These findings highlight that enforcing the Gibbs phase rule during training provides the most effective guidance toward physically consistent predictions without compromising model stability.

**Table 3. Loss-weight sweep for physics-informed training. Overall Macro-F1 on the held-out validation split after 20 epochs with fixed hyperparameters.**

|  | Weight ($\lambda_q$) | | | | | | | | |
|---|---|---|---|---|---|---|---|---|---|
|  | **0.05** | **0.10** | **0.15** | **0.20** | **0.25** | **0.30** | **0.35** | **0.40** | **0.45** |
| **Gibbs phase rule** | 0.9367 | 0.9370 | 0.9382 | 0.9360 | 0.9352 | 0.9334 | 0.9347 | 0.9348 | 0.9327 |
| **Local smoothness** | 0.9344 | 0.9285 | 0.9299 | 0.929 | 0.9282 | 0.9273 | 0.9264 | 0.9236 | 0.9261 |
| **Pure phase feasibility** | 0.9335 | 0.9335 | 0.9303 | 0.9300 | 0.9283 | 0.9281 | 0.9249 | 0.9265 | 0.9266 |

Beyond the physics-informed loss, we also evaluated the mechanics of the hybrid projection (pure phase feasibility → local smoothness → Gibbs phase rule cap) applied to the seed ensemble probabilities. As established in Section 3.1, this projection yielded the most stable and accurate predictions across systems (Table 2 and Figure 6). Our ablations confirmed that the operator order in this decoding step was critical for these gains. Applying pure phase feasibility first eliminated corner violations; a local-smoothness filter then attenuated low-confidence, spatially isolated activations by averaging probabilities within a small composition-temperature neighborhood; finally, the Gibbs cap resolved any remaining over-prediction of phase multiplicity. The improvements concentrated where phase boundaries were dense, e.g., Bi-Sn showed clearer improvements (Macro-F1 0.975; Subset Accuracy 94.5%), while near-ceiling systems saw little change (e.g., Ag-Bi-Cu). On already high-accuracy systems, projection rendered performance essentially unchanged. Seed ensembling and projection were complementary: averaging per-phase probabilities across seeds reduces stochastic noise, and the subsequent projection removed residual unphysical labels, jointly tightening confidence and stability. In short, relative to the base GNN, projection provided modest mean gains, substantial variance reductions, and physically consistent predictions across both binary and ternary evaluations.

Figure 7 qualitatively compares phase-multiplicity maps for two representative binary systems, (a) Ag-Bi and (b) Bi-Sn, across GNN, GNN+Physics+Informed Loss, and GNN+Physics-Informed Decoding. In the baseline GNN, physically inconsistent multiplicities appeared in two characteristic ways. First, within the binary composition interior, the model occasionally predicted three-phase coexistence, which exceeded the Gibbs phase rule cap for a binary system. Second, near the pure phase corners/edges, the GNN sometimes produced multi-phase outputs, which were likewise inconsistent with the Gibbs phase rule constraint that collapsed feasible multiplicity to a single phase in these regions. When trained with the Gibbs phase rule physics-informed loss, the frequency and spatial extent of these violations were substantially reduced: spurious three-phase islands in binaries became rarer, and corner/edge regions more often reverted to single-phase predictions, consistent with the Gibbs phase rule-induced cap. Importantly, because the Gibbs phase rule term was enforced as a soft penalty during optimization, occasional localized violations could still remain even at $\lambda_{GPR} = 0.15$.

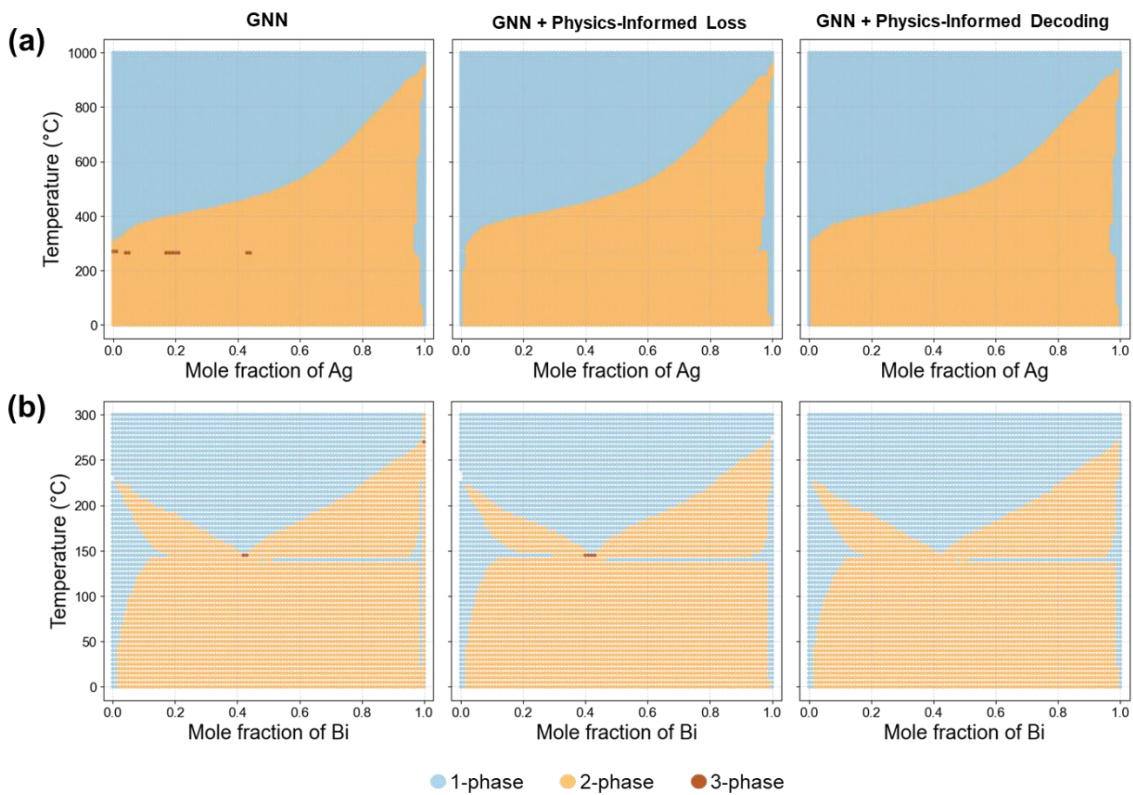

**Figure 7. Phase-multiplicity maps over composition-temperature space for two binary systems: (a) Ag-Bi and (b) Bi-Sn. Columns compare GNN, GNN+Physics-Informed Loss, and GNN+Physics-Informed Decoding. Colors denote predicted phase multiplicity (1, 2, or 3 phases). The baseline GNN showed infeasible outputs, including three-phase predictions in binaries and multi-phase predictions at pure phase corners. Physics-informed loss reduced these violations but did not guarantee complete removal, whereas decoding enforced feasibility deterministically and eliminated violations by construction.**

This contrasts with physics-informed decoding, which applied feasibility enforcement deterministically and therefore removed Gibbs phase rule violations by construction. The benefit of physics-based decoding is also visually apparent in Figure 7. While the physics-informed loss improved physical consistency in expectation, it could not guarantee feasibility everywhere because training must balance the supervised objective against a soft constraint term. Decoding, on the other hand, applied a deterministic feasibility projection to the predicted label sets, ensuring that physically inadmissible multiplicities are never returned. As a result, binary predictions contained no spurious three-phase regions, and pure phase corners no longer exhibited multi-phase outputs. This distinction explains why decoding can reliably eliminate constraint violations even in boundary-dense subsystems, whereas the loss-based approach primarily reduces such errors.

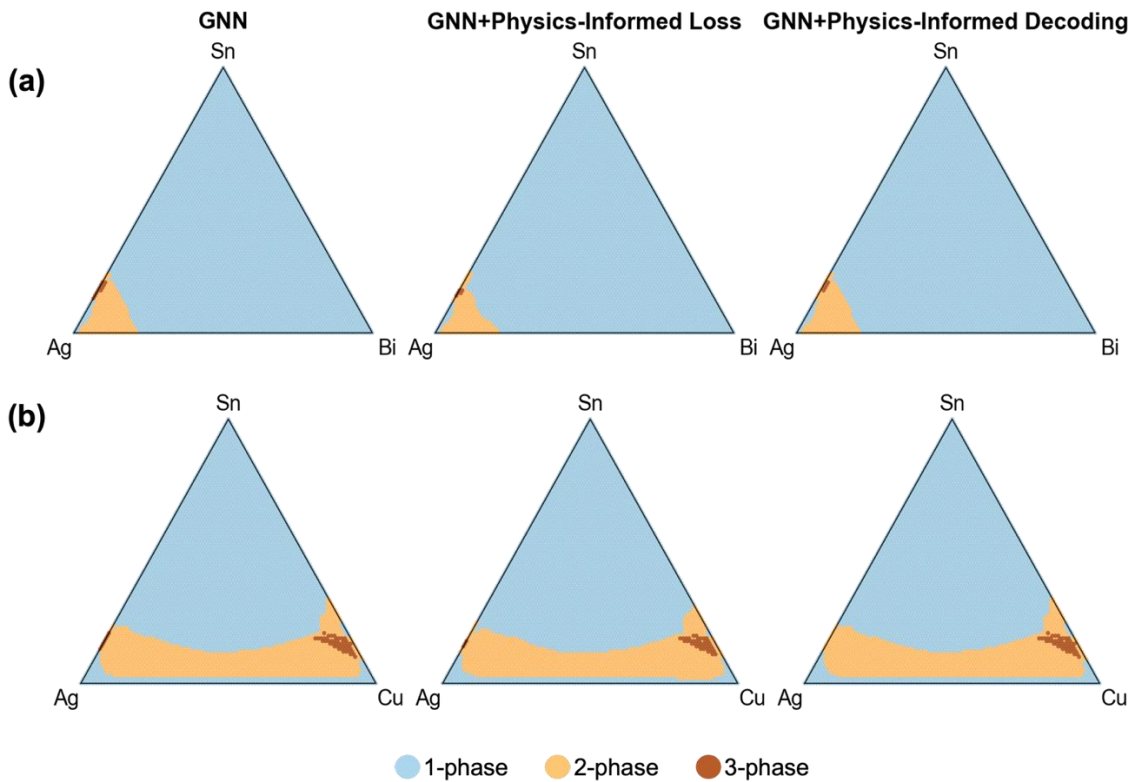

**Figure 8. Phase-multiplicity maps over composition-temperature space for two ternary systems. Rows correspond to the (a) Ag-Bi-Sn and (b) Ag-Cu-Sn systems. Columns compare GNN, GNN+Physics-Informed Loss, and GNN+Physics-Informed Decoding. Colors denote predicted phase multiplicity (1,2, or 3 phases).**

Building upon the binary results, similar and equally critical improvements were observed when evaluating the higher-dimensional ternary subsystems. Figure 8 visually illustrates the phase-multiplicity maps for two representative ternary systems across the composition space. These maps categorize every composition point by its predicted phase count, offering a direct visualization of the models' thermodynamic compliance. For the unconstrained baseline GNN, physically inconsistent predictions occasionally emerged. Because the purely data-driven model lacks an inherent understanding of thermodynamic bounds, diffuse probabilities near complex phase boundaries or pure phase corners frequently lead to spurious multiphase regions. In a ternary isothermal-isobaric section, the Gibbs phase rule dictated a strict cap of three coexisting phases; however, the baseline model was susceptible to over-predicting this multiplicity, yielding chemically impossible phase assemblages.

The incorporation of a physics-informed loss function actively penalized these violations during training, guiding the learning process toward more physically consistent solutions. While this soft regularization approach substantially reduced the frequency and spatial extent of unphysical phase counts, it must balance the physical constraint against the primary supervised classification loss. Consequently, isolated localized violations could still persist even in the fully converged model. By contrast, the implementation of physics-informed decoding bypassed the issue of competing gradients entirely. The application of a deterministic post-hoc projection, this method explicitly pruned impossible phase combinations at the elemental corners and strictly enforced the global cardinality bound dictated by the Gibbs phase rule. As visually evident in the decoded maps in Figure 8, the projections were completely free of multiplicity violations. This guaranteed that the final predicted phase assemblages remain strictly physically admissible across the entire ternary domain, fundamentally improving the reliability of the surrogate model for downstream metallurgical interpretation.

### 3.3 In-domain and out-of-domain dense-grid predictions

To assess interpolation quality within the training domain, we first evaluated the final model on an in-domain dense composition-temperature mesh for all six binaries (Ag-Bi, Ag-Cu, Ag-Sn, Bi-Cu, Bi-Sn, Cu-Sn) and three ternaries (Ag-Bi-Cu, Ag-Cu-Sn, Bi-Cu-Sn). We then tested out-of-domain extrapolation on the Ag-Bi-Sn ternary and the Ag-Bi-Cu-Sn quaternary, which were outside the training compositions, to evaluate the model's ability to generalize to previously unseen composition spaces. We restricted this evaluation to physics-informed decoding, as it demonstrated superior performance compared to the physics-informed loss approach, as discussed in Sections 3.1 and 3.2.

Evaluating in-domain and out-of-domain dense-grid predictions is critical because it directly tests the models' ability to interpolate reliably within the space it was trained on. We probed and stress-tested the models' fidelity at resolutions beyond the coarser learning grid by generating a dense composition-temperature mesh at 1 at.% composition spacing and 5 °C temperature spacing. This sampling was substantially finer than the original training data. This allowed us to identify subtle errors, quantify the mismatch count, and compute exact-set accuracy, thereby assessing how well the models could generalize to intermediate compositions and temperatures within the training domain. Such evaluation is essential for practical applications, as material design often requires high-resolution predictions in composition-temperature space to identify stable phases, guide experiments, and ensure confidence in predictions for compositions that were not explicitly part of the training set. All predictions were made using the seed ensemble probabilities. The performance on these dense grids is summarized by the mismatch count and the associated exact-set accuracy, as defined in Section 2.4.

Table 4 summarizes the performance of the GNN with physics-informed decoding across both interpolation and extrapolation regimes. Focusing first on the interpolation results (six binaries and three in-domain ternaries), the model demonstrated consistently high exact-set accuracy despite the substantially increased evaluation resolution. Among the binaries, Bi-Cu achieved the highest accuracy (97.34%), while Ag-Sn and Cu-Sn were slightly lower (95.94%), reflecting the sharper phase boundary curvature and narrower transition regions in these systems, which made predictions more sensitive to small composition-temperature variations. These results indicate that errors are not widespread across phase-field interiors but are predominantly localized near phase boundaries, where the equilibrium phase set can change abruptly on the fine grid.

For the three in-domain ternaries, dense-grid accuracy remained comparably high at 96.18 ± 0.06%, ranging from 96.13% to 96.27%, showing that the model preserved its interpolation quality when transitioning from simpler binary to more complex ternary phase-field topologies. Averaged over the six binaries, the dense-grid exact-set accuracy was 96.36 ± 0.47% (mean ± std), spanning 95.94% to 97.34%. These high accuracies on fine composition-temperature meshes demonstrate the model's ability to reliably interpolate within the training domain, capturing the correct phase behavior across both broad regions and intricate phase boundaries.

**Table 4. Interpolation and extrapolation performance of GNN with physics-informed decoding.**

| | Interpolation | | | | | | | | | Extrapolation | |
|---|---|---|---|---|---|---|---|---|---|---|---|
| | Binary | | | | | | Ternary | | | Ternary | Quaternary |
| | Ag-Bi | Ag-Cu | Ag-Sn | Bi-Cu | Bi-Sn | Cu-Sn | Ag-Bi-Cu | Ag-Cu-Sn | Bi-Cu-Sn | Ag-Bi-Sn | Ag-Bi-Cu-Sn |
| Exact-set match count | 20351/21129 | 20347/21129 | 20295/21153 | 20583/21149 | 6167/6402 | 20297/21155 | 5459/5679 | 5524/5738 | 5369/5584 | 5581/5619 | 26487/28860 |
| Accuracy [%] | 96.3189 | 96.2989 | 95.9438 | 97.3379 | 96.3293 | 95.9442 | 96.1261 | 96.27 | 96.1497 | 99.32 | 91.7775 |

We assessed out-of-domain generalization on two settings that were entirely excluded from training: (i) the Ag-Bi-Sn ternary at an isothermal plane of 700 °C sampled on a 1 at.% mesh, and (ii) the Ag-Bi-Cu-Sn quaternary at 700 °C sampled on a 2 at.% tetrahedral mesh. On the Ag-Bi-Sn plane, the model achieved an impressive 99.32% exact-set accuracy, with only 38 mismatches out of 5,619 evaluated points (Table 4). Much of this plane was dominated by a single-phase LIQUID region at 700 °C, which naturally raised the attainable accuracy; nonetheless, residual errors were concentrated in eutectic and peritectic regions, as well as multi-phase tie triangles, where small compositional changes could lead to abrupt shifts in the equilibrium phase set. For the Ag-Bi-Cu-Sn quaternary, despite being a higher-dimensional surface entirely absent from training, the model still attained a strong 91.78% exact-set accuracy, with 2,373 mismatches out of 28,860 data points. While the accuracy was lower than for simpler systems, achieving over 90% on a previously unseen quaternary demonstrates the model's robust generalization to higher-order, complex composition spaces. Moreover, the pattern of errors was localized near sharp phase boundaries, suggesting that the model reliably captures the bulk of the phase-field behavior even in extrapolated regions. These results highlight that the physics-informed decoding GNN can not only interpolate accurately within the training domain but also provide trustworthy predictions in out-of-domain systems, making it a valuable tool for guiding experimental exploration and computational design in multi-component materials.

Figure 9 provides a qualitative, high-resolution diagnostic of in-domain interpolation on representative binaries by pairing an exact-set match/mismatch map (left) with the corresponding predicted phase-set map (right). Across panels (a)-(c), a consistent pattern emerges. Mismatches are sparse within the interiors of stable phase fields and concentrate primarily along narrow, structured bands that trace phase-boundary neighborhoods. Specifically, in (a) Ag-Bi, mismatch points form a thin curved band that follows the principal transition separating broad stability regions. In (b) Ag-Cu, mismatches are tightly localized near the high-temperature transition band. In (c) Ag-Sn, the mismatch pattern similarly aligns with the dominant boundary curves and localized transition neighborhoods, reinforcing that dense-grid errors concentrate where the phase set is most sensitive to small composition-temperature perturbations.

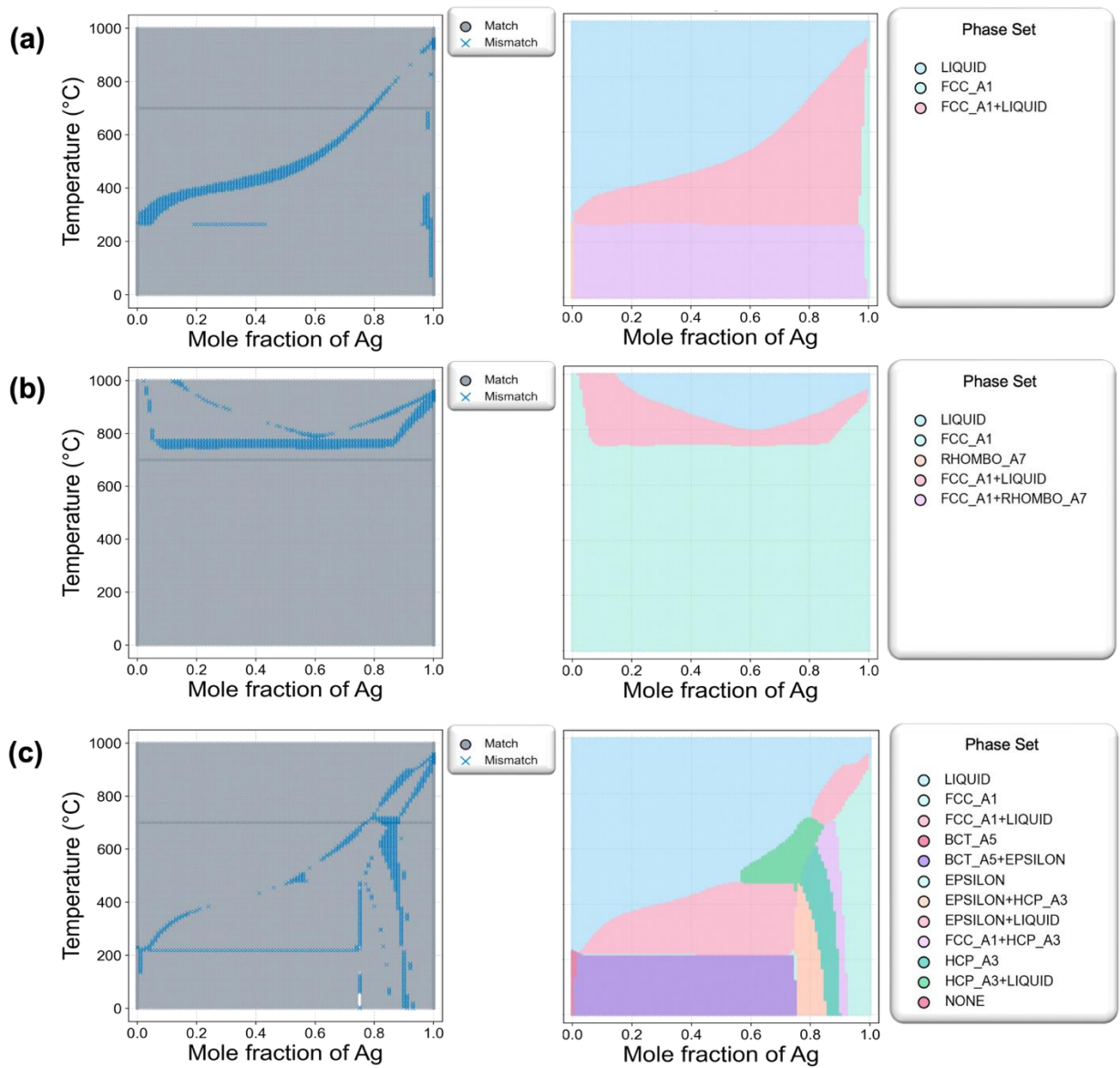

**Figure 9. In-domain dense-grid (1 at.% and 5 °C) binary predictions. Left: match/mismatch map, Right: predicted phase map. (a) Ag-Bi, (b) Ag-Cu, (c) Ag-Sn.**

Figure 10 extends this high-resolution diagnostic to the remaining in-domain binary systems: (a) Bi-Cu, (b) Bi-Sn, and (c) Cu-Sn. Consistent with the previously discussed binaries, the predicted phase-set maps maintain spatial coherence with well-defined interfaces. Mismatches are notably concentrated near complex invariant-reaction neighborhoods, such as eutectic and peritectic regions, where multiple phase sets closely compete. The broad single-phase and two-phase regions exhibit near-uniform agreement with CALPHAD labels, confirming that the model captures the underlying thermodynamic stability with high fidelity even at an exceptionally fine grid resolution.

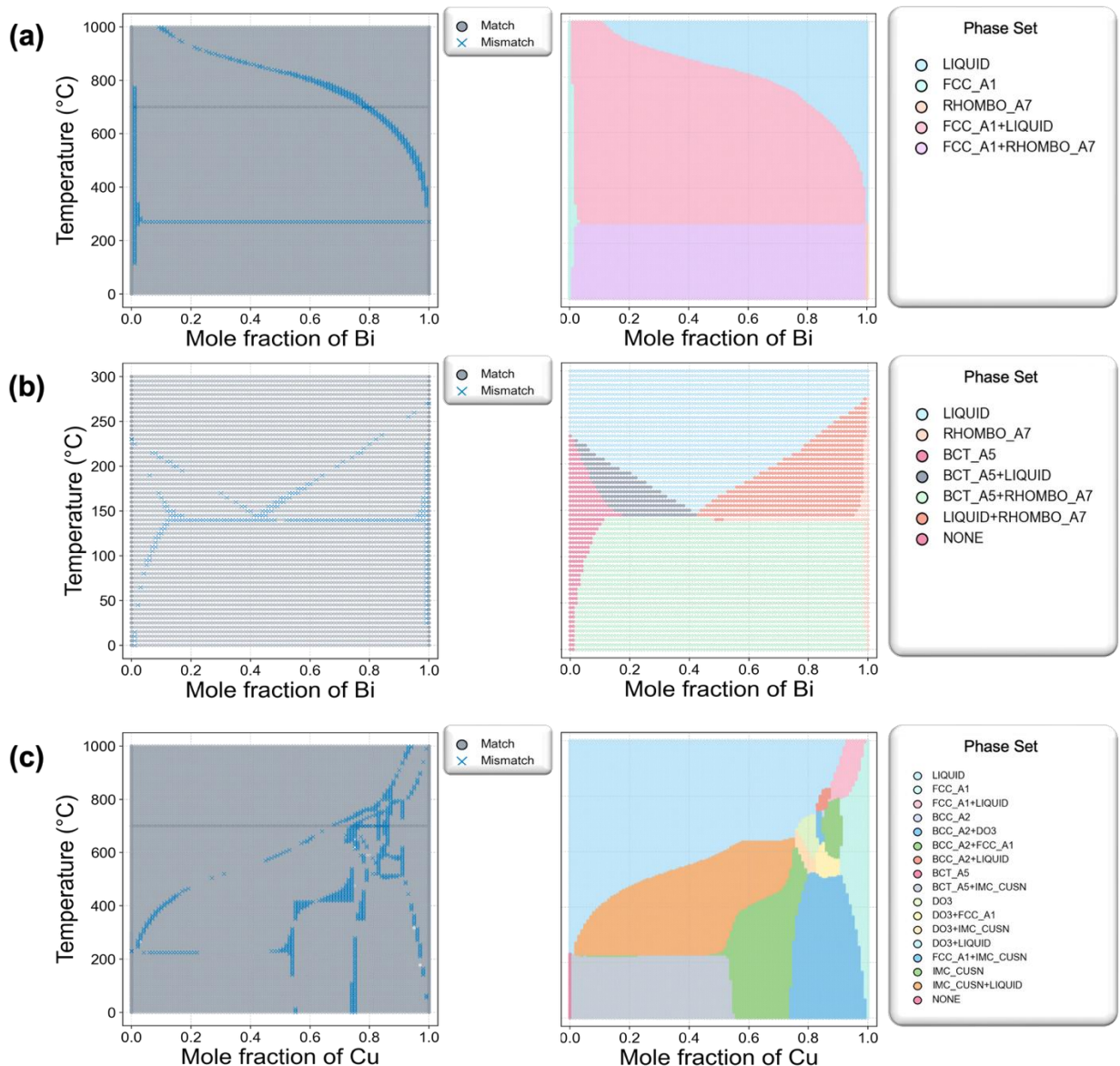

**Figure 10. In-domain dense-grid (1 at.% and 5 °C) binary predictions. Left: match/mismatch map, Right: predicted phase map. (a) Bi-Cu, (b) Bi-Sn, (c) Cu-Sn.**

Figure 11 compares dense-grid predictions for the ternary systems at an isothermal section of 700 °C for both in-domain and out-of-domain cases. The left panels display exact-set match/mismatch maps against CALPHAD labels, while the right panels show the corresponding predicted phase maps. Panels (a) Ag-Cu-Sn, (b) Ag-Bi-Cu, and (c) Bi-Cu-Sn demonstrate the model's in-domain interpolation capabilities. In these systems, mismatches remain sparse in phase-field interiors and are localized along narrow bands near phase boundaries, reflecting the model's robust interpolation quality. Furthermore, panel (d) highlights the out-of-domain extrapolation performance on the completely unseen Ag-Bi-Sn system. Despite the lack of direct training data for this ternary space, the predictions remain highly accurate, with minor residual errors confined primarily to complex multiphase regions.

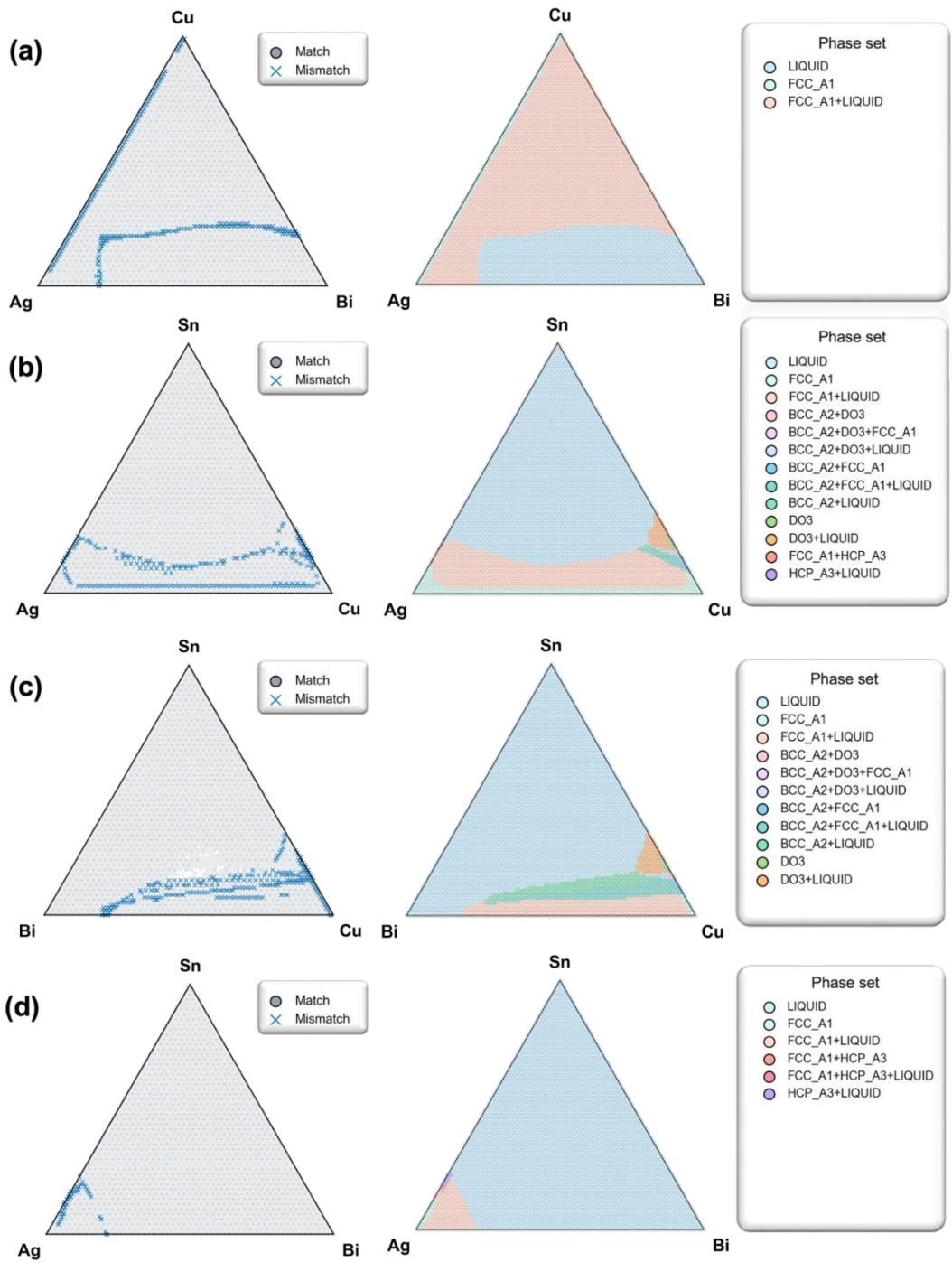

**Figure 11. Dense-grid ternary predictions at 700 °C. Left: match/mismatch map, Right: predicted phase map. In-domain interpolation results are shown for (a) Ag-Cu-Sn, (b) Ag-Bi-Cu, (C) Bi-Cu-Sn). Out-of-domain extrapolation: (d) Ag-Bi-Sn.**

To assess generalization to higher-order dimensions, Figure 12 illustrates the out-of-domain dense-grid prediction for the unseen Ag-Bi-Cu-Sn quaternary system at 700 °C. The left panel visualizes the match/mismatch distribution within the tetrahedral composition space, and the right panel presents the corresponding predicted phase-set map. Most mismatches lie on or near the complex phase-change surfaces of the tetrahedron, where minute atomic-fraction changes can push a state across a multiphase-region boundary. Nevertheless, the hybrid projection effectively maintains physical consistency in this extrapolated regime: the pure phase rule prevents spurious intermetallic activations near vertices, and local smoothness preserves sharp structural transitions. These results indicate that the physics-informed graph representation supports meaningful and stable extrapolation to complex quaternary surfaces.

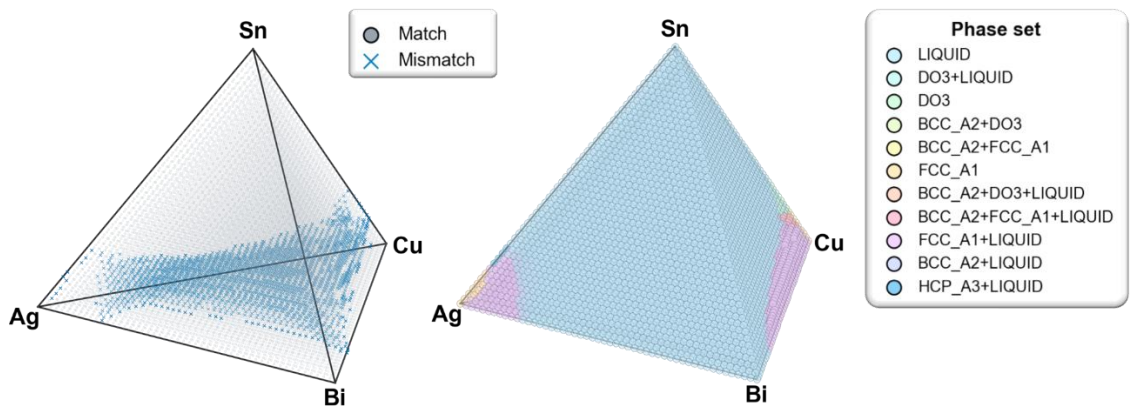

**Figure 12.** Out-of-domain (extrapolation) dense-grid predictions for the Ag-Bi-Cu-Sn quaternary system at 700 °C. Left: match/mismatch map, Right: predicted phase map.

## 4. Conclusions

We treated phase-diagram inference as a multi-label phase-set prediction problem over composition-temperature space and introduced a graph neural network augmented with physics-informed loss and a physics-informed decoder. A deliberate separation of physics-informed training and post hoc decoding proved crucial. A single physics loss stabilized optimization without diluting supervision, while the deterministic decoder enforced feasibility at inference, allowing multiple constraints to be applied without gradient conflicts. This separation contributed to observed stability gains and sharper boundaries in phase-dense regions. Together, they produced stable, physically faithful phase maps from which emergent features, such as triple points, could be inferred directly from composition-temperature data.

Trained on six binaries and three ternaries in Ag-Bi-Cu-Sn, the models achieved high accuracy on dense in-domain grids and generalized effectively to an unseen ternary (Ag-Bi-Sn) and a quaternary surface (Ag-Bi-Cu-Sn at 700 °C), demonstrating strong exact-set match performance. Across systems, physics-informed strategies provided modest gains in mean accuracy but significantly reduced variance, yielding predictions that were both accurate and physically consistent. Even in out-of-domain settings, the model achieved >90% exact-set accuracy, highlighting robust generalization and reliability for unseen ternary and quaternary systems.

This approach is alloy-agnostic, naturally extensible to higher-component systems, and compatible with emerging large-scale thermodynamic datasets. Future work will incorporate richer thermodynamic descriptors, e.g., phase-specific Gibbs energies, lever-rule constraints, and activity coefficients to further strengthen mechanistic consistency and enable differentiable fraction predictions. Scaling to larger element sets and aggregating open datasets promises a practical surrogate for multi-component phase diagrams, unifying graph-based learning with lightweight, rigorous thermodynamic constraints.

**CRediT authorship contribution statement**
Eunjeong Park: Data curation, Formal analysis, Investigation, Methodology, Software, Validation, Visualization, Writing – original draft, Writing – review & editing. Amrita Basak: Writing – review & editing, Supervision, Resources, Project administration, Funding acquisition, Conceptualization.

**Funding information**
This work was supported by the University Graduate Fellowship at Penn State and by the Office of Naval Research under Grant No. N00014-25-1-2250. The opinions, findings, and conclusions expressed herein are solely those of the authors and do not necessarily represent the views of the funding agencies.

**Declaration of Generative AI and AI-assisted technologies in the writing process**
During this work's preparation, OpenAI ChatGPT has been used to improve grammar and readability.

**Declaration of competing interest**
The authors declare that they have no known competing financial interests or personal relationships that could have appeared to influence the work reported in this paper.

**Data availability**
Data will be made available upon reasonable request.